%% file: neurips_2025.tex
\documentclass{article}

\usepackage[dandb]{neurips_2025}

\usepackage[utf8]{inputenc} 
\usepackage[T1]{fontenc}    
\usepackage{hyperref}       
\usepackage{url}            
\usepackage{booktabs}       
\usepackage{amsfonts}       
\usepackage{nicefrac}       
\usepackage{microtype}      
\usepackage{xcolor}         
\usepackage{authblk}        
\usepackage{enumitem}
\usepackage{array} 
\usepackage{amsmath}
\usepackage{tabularx}
\usepackage{algorithm}
\usepackage{algpseudocode}
\usepackage{titletoc}
\usepackage{amssymb}
\usepackage{multirow}
\usepackage{tcolorbox}
\usepackage{parskip}
\usepackage{pifont}
\usepackage{cleveref}
\usepackage{booktabs} 
\usepackage{array} 
\usepackage{amsmath,bm}

\definecolor{my_green}{RGB}{51,102,0}
\definecolor{my_red}{RGB}{204, 0, 0}
\definecolor{my_half}{RGB}{226,115,0} 
\newcommand{\halfcheck}{\textcolor{my_half}{\ding{51}\rotatebox[origin=c]{-9.2}{\kern-0.7em\ding{55}}}}
\definecolor{paired-light-blue}{RGB}{198, 219, 239}
\definecolor{paired-dark-blue}{RGB}{49, 130, 188}
\definecolor{paired-light-orange}{RGB}{251, 208, 162}
\definecolor{paired-dark-orange}{RGB}{230, 85, 12}
\definecolor{paired-light-green}{RGB}{199, 233, 193}
\definecolor{paired-dark-green}{RGB}{49, 163, 83}
\definecolor{paired-light-purple}{RGB}{218, 218, 235}
\definecolor{paired-dark-purple}{RGB}{117, 107, 176}
\definecolor{paired-light-gray}{RGB}{217, 217, 217}
\definecolor{paired-dark-gray}{RGB}{99, 99, 99}
\definecolor{paired-light-pink}{RGB}{222, 158, 214}
\definecolor{paired-dark-pink}{RGB}{123, 65, 115}
\definecolor{paired-light-red}{RGB}{231, 150, 156}
\definecolor{paired-dark-red}{RGB}{131, 60, 56}
\definecolor{paired-light-yellow}{RGB}{231, 204, 149}
\definecolor{paired-dark-yellow}{RGB}{141, 109, 49}  
\definecolor{myblue}{RGB}{218,232,252}
\definecolor{mygray}{RGB}{220,220,220}
\definecolor{mypink}{RGB}{251,49,153}

\usepackage{xspace}

\usepackage{caption}
\usepackage{makecell}
\usepackage{colortbl}
\usepackage{enumitem}
\usepackage{caption}
\usepackage{graphicx}
\usepackage{float} 
\usepackage{wrapfig}
\usepackage{subcaption}
\usepackage{hyperref}

\usepackage{graphicx}

\title{TextSculptor: Training and Benchmarking \\ Scene Text Editing}

\author{%
  \textbf{Yiheng Lin}\textsuperscript{1,2,*},
  \textbf{Siyu Jiao}\textsuperscript{1,2,*},
  \textbf{Xiaohan Lan}\textsuperscript{2*},
  \textbf{Wei Zhou}\textsuperscript{2},
  \textbf{Qi She}\textsuperscript{2},
  \textbf{Fei Yu}\textsuperscript{2},
  \textbf{Heyun Chen}\textsuperscript{2},
  \\
  \textbf{Zhengwei Wang}\textsuperscript{2},
  \textbf{Jinghuan Chen}\textsuperscript{2},
  \textbf{Moran Li}\textsuperscript{2},
  \textbf{Yingchen Yu}\textsuperscript{2},
  \textbf{Zijian Feng}\textsuperscript{2},
  \textbf{Yao Zhao}\textsuperscript{1},
  \\
  \textbf{Yunchao Wei}\textsuperscript{1$^\dagger$},
  \textbf{Yujie Zhong}\textsuperscript{2$^\dagger$}
}

\affil{
  {\tt 
  $*$ Equal Contributors, $\dagger$ Corresponding Authors
  }
  \par
  \textsuperscript{1} Beijing Jiaotong University, 
  \textsuperscript{2} Bytedance
  \\
}

\begin{document}

\maketitle
\input{sec/0_abstract}
\input{sec/1_introduction}
\input{sec/2_related_work}

\input{sec/3_datapipeline}

\input{sec/4_benchmark}

\input{sec/5_evaluation}

\input{sec/6_conclusion}

\bibliographystyle{plain}
\bibliography{ref}

\end{document}

%% file: sec/0_abstract.tex
\begin{abstract}
Recent advances in Multimodal Large Language Models (MLLMs) and diffusion-based generative models have substantially improved prompt-driven image editing. However, scene text editing remains challenging, as it requires models to precisely modify textual content while preserving visual realism and non-target regions. Current open-source models still lag behind proprietary systems, largely due to the scarcity of high-quality training data and the lack of standardized benchmarks tailored to text editing. To address these challenges, we present \textbf{TextSculptor}, a comprehensive framework for data construction and evaluation of scene text editing. We first develop an automated data construction pipeline that combines text-aware image synthesis with programmatic text rendering and compositing. Based on this pipeline, we build \textbf{TextSculpt-Data}, a large-scale dataset containing \textbf{3.2M} training samples, including \textbf{1.2M} OCR-verified text-to-image samples and \textbf{2M} paired text editing samples with naturally aligned source--target images and strong background consistency. We further introduce \textbf{TextSculpt-Bench}, a benchmark covering four fundamental text editing tasks: text addition, text replacement, text removal, and hybrid editing. To support reliable evaluation, we design a tailored protocol that measures text accuracy, visual quality, and background preservation through OCR-based text alignment, multimodal judgment, and background-region similarity. Extensive experiments show that TextSculptor improves open-source text editing performance and narrows the gap to proprietary models. The data and benchmark are available at \href{https://github.com/linyiheng123/TextSculptor}{https://github.com/linyiheng123/TextSculptor}.

\end{abstract}

%% file: sec/1_introduction.tex
\section{Introduction}\label{sec: introduction}

\input{tex/teaser}

Recent years have witnessed rapid progress in image generation and editing, propelled by large-scale, high-quality open-source datasets~\cite{ye2025imgedit, wang2025gpt, qian2025pico, ye2025echo} and the tight integration of Multimodal Large Language Models (MLLMs) with diffusion-based generative models~\cite{wu2025qwen, team2026firered, wu2025omnigen2, jiao2025thinkgen}. Among these advances, \emph{prompt-driven image editing} has emerged as a particularly compelling paradigm: it enables users to modify image content with natural language instructions, substantially lowering the barrier to high-quality content creation. Encouragingly, open-source models such as Qwen-Image~\cite{wu2025qwen} and LongCat-Image~\cite{LongCat-Image} have demonstrated strong general editing performance, narrowing the gap to leading closed-source systems including Gemini-3-Pro-Image~\cite{google2025nanobananapro}, GPT-Image~\cite{openAI2025gpt4o}, and Seedream~\cite{seedream2025seedream}. Together, these developments signal the maturation and democratization of high-fidelity, semantically grounded image editing.

As general editing capabilities improve, recent studies \cite{wu2025qwen, LongCat-Image, team2025zimage} have increasingly focused on the more challenging setting of \emph{scene text} rendering and editing. Nevertheless, a clear performance gap remains between current open-source models and state-of-the-art closed-source models on text editing tasks. We attribute this gap to two fundamental bottlenecks: (\textit{i}) the scarcity of high-quality open-source training data for text editing, and (\textit{ii}) the lack of precise and standardized evaluation benchmarks. On the data side, most existing open-source resources \cite{chen2023textdiffuser, tuo2023anytext, wang2025textatlas5m, li2024densefusion} are designed primarily for \emph{text-to-image} generation rather than instruction-following text editing. On the evaluation side, existing benchmarks \cite{du2025textcrafter, geng2025x} mainly assess text generation quality; while some general editing benchmarks \cite{liu2025step1x} include a limited number of text-editing cases, they do not provide comprehensive coverage of the diverse failure modes specific to text editing.

To address the data bottleneck, we propose an automated construction pipeline and build a high-quality dataset for both text rendering and text editing, dubbed \textbf{TextSculpt-Data}. The dataset consists of two complementary parts. The first part targets \emph{text-to-image} to strengthen fundamental text rendering ability. Specifically, we use Qwen3-VL to rewrite captions from the open-source dataset \cite{gadre2023datacomp}, by inserting contextually appropriate text descriptions and placement cues. We then leverage strong image generation models~\cite{wu2025qwen, seedream2025seedream, google2025nanobananapro} to synthesize high-resolution images, and apply an OCR-based quality gate to filter low-legibility samples using word-level accuracy. 
The second part targets \emph{text editing}. We sample high-frequency vocabulary, combine it with diverse open-source fonts, render paired text layers with Python-based engines, and composite them onto natural images to form aligned source--target editing pairs. This programmatic synthesis not only provides exact text ground truth, but also encourages strong background preservation by construction, since non-edited pixels remain unchanged outside the edited regions.
Overall, the resulting pipeline generates more than 3M text-to-image candidates, from which \textbf{1.2M} OCR-verified samples are retained. Together with \textbf{2M} paired editing samples, TextSculpt-Data contains \textbf{3.2M} training samples in total.

To validate the effectiveness of our data and address the lack of standardized evaluation for text image editing, we further introduce \textbf{TextSculpt-Bench}, a benchmark designed to assess the fundamental capabilities of models in text editing scenarios. As shown in Figure~\ref{fig:teaser}, TextSculpt-Bench covers four essential task types: \emph{text addition}, \emph{text replacement}, \emph{text removal}, and \emph{hybrid editing}, which together capture the core abilities required for text image editing. Our benchmark evaluates models along three key dimensions: \emph{text accuracy}, \emph{visual quality}, and \emph{background preservation}. Unlike previous benchmarks~\cite{gui2025texteditbench, zhang2026weedit}, which rely on VLMs to score all dimensions, our evaluation is tailored to text editing by combining multimodal judgment with OCR-based measurement. Specifically, text accuracy is measured via whole-image text alignment with word-level edit errors, visual quality is assessed through location correctness, style consistency, and physical plausibility, and background preservation is computed as SSIM on OCR-masked background regions. This design enables more reliable and fine-grained evaluation of text editing performance.

Our main contributions are summarized as follows:
\begin{itemize}[itemsep=2pt,topsep=0pt,parsep=0pt]
\item \textbf{Automated data construction pipeline.} We propose an automated pipeline for large-scale text image data construction, combining VLM-based caption rewriting, high-quality image synthesis, and programmatic text compositing. This pipeline enables scalable generation of training data for both text rendering and text editing.

\item \textbf{Large-scale text image dataset.} We build \textbf{TextSculpt-Data}, a high-quality dataset consisting of \textbf{1.2M} text-to-image samples and \textbf{2M} text editing pairs. The dataset provides strong supervision for faithful text rendering and precise text editing, while naturally encouraging background preservation through aligned construction.

\item \textbf{Benchmark for fundamental text editing.} We introduce \textbf{TextSculpt-Bench}, a benchmark targeting the fundamental capabilities of text image editing through four essential task types: text addition, text replacement, text removal, and hybrid editing. The benchmark evaluates models along three dimensions: text accuracy, visual quality, and background preservation, using a tailored protocol that combines multimodal judgment with OCR-based measurement.
\end{itemize}

%% file: tex/teaser.tex
\begin{figure*}[t]
\vspace{-6pt}
\begin{center}
   \includegraphics[width=0.98\linewidth]{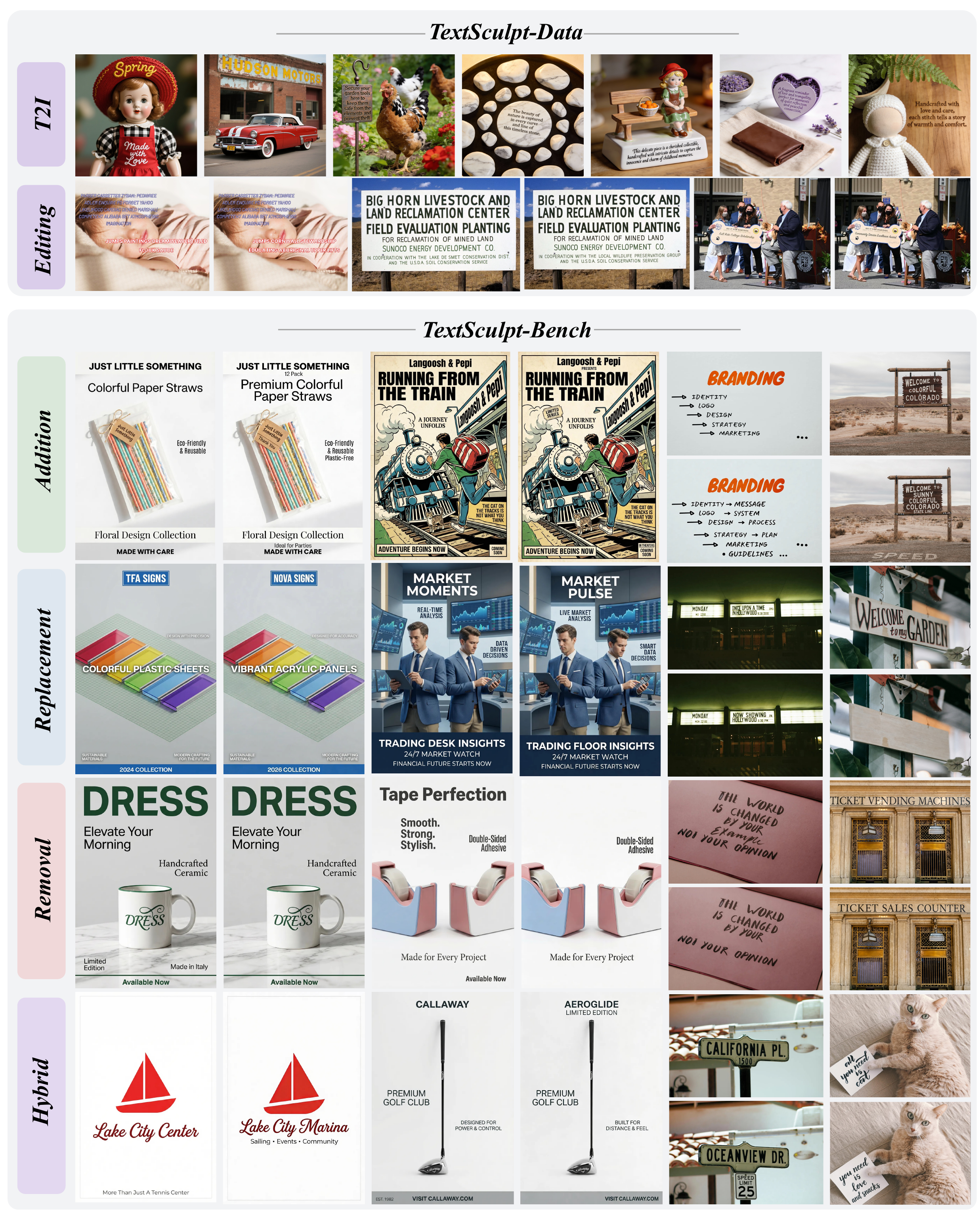}
\end{center}
   \caption{
    Illustration of TextSculpt-Data and TextSculpt-Bench. TextSculpt-Data contains text-to-image rendering data and paired editing data, while TextSculpt-Bench evaluates four task types: addition, replacement, removal, and hybrid editing.
   }
\label{fig:teaser}
\end{figure*}

%% file: sec/2_related_work.tex
\section{Related Work}

\subsection{Instruction-Guided Image Editing}

Instruction-guided image editing aims to modify visual content according to natural language instructions while preserving irrelevant regions. Early diffusion-based methods usually formulate editing as conditional generation, where the input image and text instruction are jointly encoded to guide local or global image modification. Representative works such as InstructPix2Pix~\cite{brooks2023instructpix2pix} and InstructEdit~\cite{wang2023instructeditimprovingautomaticmasks} establish practical pipelines for instruction-following image editing, while MagicBrush~\cite{zhang2024magicbrushmanuallyannotateddataset} provides manually annotated real-image editing triplets for training and evaluation. Later methods improve editing controllability by introducing larger instruction datasets, stronger multimodal encoders, or more explicit region-level guidance. For example, AnyEdit~\cite{anyedit}, UltraEdit~\cite{zhao2024ultraeditinstructionbasedfinegrainedimage}, SmartEdit~\cite{huang2023smarteditexploringcomplexinstructionbased}, and Step1X-Edit~\cite{liu2025step1x} extend instruction-based editing to more diverse object, style, and layout modifications. Recent unified or generalist generative frameworks, including OmniGen~\cite{xiao2024omnigenunifiedimagegeneration}, Qwen-Image~\cite{wu2025qwen}, Bagel~\cite{deng2025emerging}, and Emu3.5~\cite{cui2025emu3}, further show the growing capability of image generation and editing within broader multimodal generation settings.

\subsection{Text-Centric Data and Evaluation}
Existing instruction-based editing datasets, such as InstructPix2Pix~\cite{brooks2023instructpix2pix}, MagicBrush~\cite{zhang2024magicbrushmanuallyannotateddataset}, HQ-Edit~\cite{hui2024hqedithighqualitydatasetinstructionbased}, UltraEdit~\cite{zhao2024ultraeditinstructionbasedfinegrainedimage}, and SEED-Data-Edit~\cite{ge2024seeddataedittechnicalreporthybrid}, mainly focus on object, attribute, or style modification, and provide limited supervision for precise text manipulation. Recent text-oriented generation and editing datasets, including TextDiffuser~\cite{chen2023textdiffuser}, AnyText~\cite{tuo2023anytext}, and TextAtlas5M~\cite{wang2025textatlas5m}, improve visual text rendering ability, but do not fully support paired instruction-following operations such as adding, replacing, removing, or jointly modifying text in existing images.

Evaluation for text editing faces a similar gap. General generation and editing benchmarks~\cite{rise, ye2025imgedit, liu2025step1x} mainly assess reasoning, compositionality, or instruction following, rather than text editing as a dedicated task. Text rendering benchmarks such as Lex-Bench~\cite{lex-bench}, TextAtlasEval~\cite{wang2025textatlas5m}, and CVTG-2K~\cite{du2025textcrafter} evaluate whether text is correctly generated in images, but they are primarily designed for text-to-image generation. Recent text editing benchmarks, such as TextEditBench~\cite{gui2025texteditbench} and WeEdit~\cite{zhang2026weedit}, move closer to text-centric editing scenarios. However, their semantic evaluation still largely relies on VLM-based scoring, which provides subjective ratings and can be insensitive to fine-grained textual errors.

%% file: sec/3_datapipeline.tex
\section{TextSculpt-Data: A High-Fidelity Text Editing Dataset}
\label{sec: Dataset}

In this section, we present the \textbf{TextSculpt-Data}, a large-scale, high-fidelity dataset specifically curated to address the scarcity of training resources for text editing. Distinct from general image editing datasets, we focus on diverse editing tasks within text-rich scenarios, prioritizing editing precision and strict background consistency. The remainder of this section is organized as follows: Section~\ref{sec:edit_type} formalizes the taxonomy of four text editing tasks supported by our dataset. Section~\ref{sec:pipeline} elaborates on our automated construction pipeline, which combines VLM-based rewriting with a programmatic rendering engine to ensure both semantic richness and ground-truth accuracy. 

\subsection{Task Definition}
\label{sec:edit_type}
We categorize text editing tasks into four fundamental operation types, covering the core capabilities required in text image editing.

\paragraph{Text Addition}
This task involves inserting new text into the image, requiring the model to synthesize plausible text layouts. We address two specific scenarios: \emph{Text-Free Surface Addition}, where text is placed onto empty carriers (e.g., blank signboards or walls). The primary challenge here is to ensure the generated text naturally adheres to the surface's curvature, perspective, and material texture. The second scenario, \emph{In-Context Insertion}, is significantly more demanding as it involves adding text amidst existing characters. Beyond geometric alignment, this requires the model to strictly match the surrounding font style and perform intelligent layout re-planning (e.g., adjusting spacing or reflowing lines) to accommodate the new content seamlessly.

\paragraph{Text Replacement}
As a fundamental operation in text image editing, replacement requires substituting existing text content while strictly preserving the original visual attributes. The model must disentangle the text content from its style (font, color, size, rotation) and the underlying background texture, ensuring that the new text blends seamlessly into the original region without creating artifacts.

\paragraph{Text Removal}
This task removes existing text from an image while restoring the underlying content coherently. We consider two common scenarios: \emph{Complete Text Erasure}, where an entire word, phrase, or sentence is removed, and \emph{In-Context Deletion}, where only a local text span within a longer expression is deleted. The former mainly tests whether the model can faithfully reconstruct the exposed background texture, illumination, and geometry without leaving visible artifacts. The latter is more challenging, as it additionally requires preserving the surrounding non-target text exactly, including its layout, spacing, and readability, while performing a localized edit.

\paragraph{Hybrid Editing}
This task requires the model to follow a composite instruction that combines text replacement, text removal, and text addition within a single image. In our benchmark, each hybrid sample contains exactly one replace target, one remove target, and one add target, requiring the model to handle multiple editing operations under a unified objective. The main challenge is to avoid cross-operation interference: replaced text should preserve the original style and placement, removed regions should be naturally inpainted, and newly added text should be integrated with plausible geometry and appearance. Successful hybrid editing therefore demands strong global consistency across the entire image.

\subsection{Data Construction Pipeline}
\label{sec:pipeline}

\input{tex/pipeline}

High-quality training data for text editing remains relatively scarce. Existing datasets~\cite{wang2025textatlas5m, tuo2023anytext} often suffer from weak image-text alignment or limited image quality, which restricts their usefulness for training text-centric generation models. To address these limitations, we develop a two-part data construction framework, as illustrated in Figure~\ref{fig:pipeline}, with each pipeline targeting a complementary capability. The first pipeline focuses on high-quality text-to-image data for building text rendering ability, while the second pipeline constructs text editing pairs through programmatic synthesis.

\subsubsection{Part 1: Text Rendering Data Pipeline}

Before tackling text editing, the model must first acquire strong text rendering ability. However, existing large-scale image-text corpora often contain text that is blurry, distorted, or structurally implausible, making them suboptimal for learning accurate text generation. To remedy this issue, we construct a text-to-image dataset that emphasizes both textual correctness and visual quality, ensuring that the synthesized text is clear, accurate, and aesthetically well integrated into the image.

We start from a large-scale image-caption corpus as the seed data. To increase the density and relevance of textual content, we use Qwen3-VL~\cite{bai2025qwen3} to rewrite the original captions. During rewriting, the model is instructed to identify plausible text-bearing surfaces in the scene (e.g., billboards, T-shirts, or mugs), generate text content that is semantically consistent with the visual context, and adapt the text length to the geometry of the target carrier.

The rewritten captions are then fed into multiple image generation models~\cite{wu2025qwen, google2025nanobananapro, seedream2025seedream} to synthesize high-quality images. To ensure strict text quality, we apply a PaddleOCR-based quality filter~\cite{cui2025paddleocr30technicalreport} and retain only samples with perfect word-level accuracy. In total, we keep 1.2M samples out of more than 3M generated candidates. This filtering step effectively removes samples with illegible or erroneous text, resulting in a high-quality dataset tailored for learning robust text rendering.

\subsubsection{Part 2: Text Editing Data Pipeline}
To obtain large-scale text editing data with precise ground truth, we develop a fully automated programmatic pipeline that balances scalability, controllability, and background fidelity.

We begin by constructing a diverse synthetic text corpus through stochastic sampling of high-frequency vocabulary\footnote{https://github.com/first20hours/google-10000-english}. To better mimic real-world text, we further introduce randomized numbers, special symbols, and casing variations.

Next, we synthesize \textbf{text editing pairs} using a Python-based rendering engine. Leveraging a large collection of open-source fonts\footnote{https://fonts.google.com/} and dynamic layout algorithms, we render the source text layer and the corresponding target text layer according to the desired editing operation. This synchronized rendering process preserves key visual attributes, such as font family, color, and stroke width, thereby providing precise ground truth for each edit.

Finally, we composite the rendered text layers onto natural images. To avoid interference with pre-existing text in images, we identify safe placement regions using OCR detection. Compared with generative inpainting, this strategy is substantially more efficient and naturally guarantees exact background preservation, since all non-edited pixels remain unchanged outside the edited regions.

In total, we construct \textbf{2M} text editing pairs using this programmatic pipeline, covering diverse editing patterns including text addition, text replacement, text removal, and hybrid editing. This corpus supports learning precise text manipulation with faithful background preservation.

%% file: tex/pipeline.tex
\begin{figure*}[t]
\begin{center}
   \includegraphics[width=0.99\linewidth]{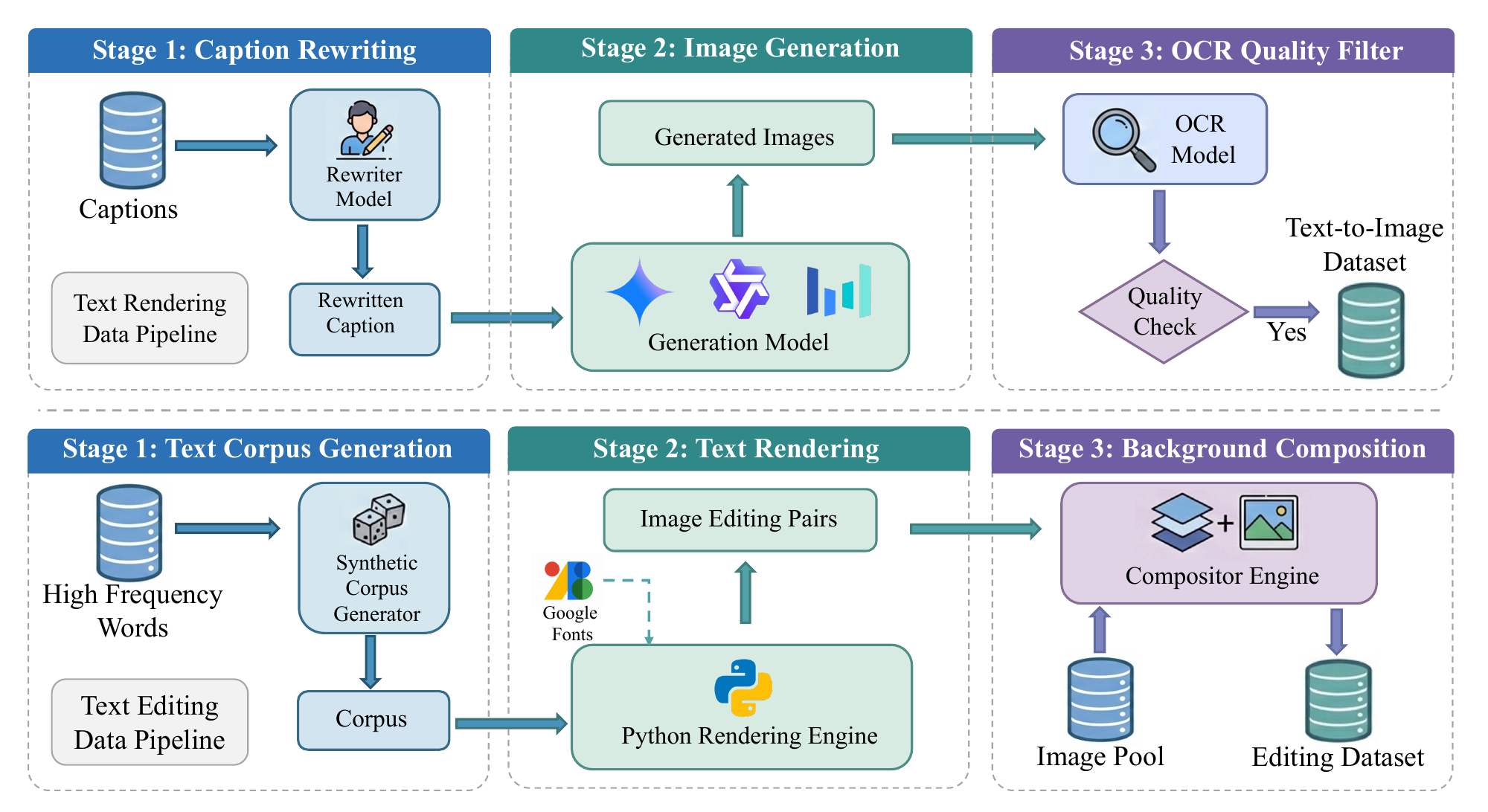}
\end{center}
   \caption{
    Illustration of our automated data construction pipelines. The top stream generates high-fidelity text-to-image samples using VLM-based rewriting and OCR filtering. The bottom stream constructs text editing pairs through a programmatic rendering and compositing engine, ensuring pixel-perfect background consistency.
   }
\label{fig:pipeline}
\end{figure*}

%% file: sec/4_benchmark.tex
\section{TextSculpt-Bench: A Comprehensive Benchmark}\label{sec:benchmark}
To facilitate a comprehensive evaluation of text editing capabilities, we introduce a novel benchmark tailored for diverse editing scenarios. Section~\ref{sec:construction} details the data collection and instruction generation pipeline, while Section~\ref{sec:metric} defines the multi-dimensional evaluation metrics.

\subsection{Benchmark Curation Pipeline}\label{sec:construction}

\paragraph{Source Image Collection}
We curate the source images from two complementary subsets. The first subset contains 188 high-quality natural images manually collected from Pexels\footnote{https://www.pexels.com/}. During collection, we exclude images that are unsuitable for text editing, such as those with overly small, sparse, or severely blurred text, to ensure sufficient visual clarity for fine-grained manipulation. The second subset consists of 168 poster-style images generated by NanoBanana. Compared with the Pexels subset, these images typically contain more numerous and denser textual elements, providing more challenging and compositionally rich scenarios for evaluating text editing models.

\paragraph{Instruction Generation}
To construct a high-quality, context-aware benchmark, we design an automated instruction generation pipeline based on the GPT-5.2~\cite{gpt5-2}. Given a source image, the model first performs visual-text analysis to identify the coherent text content in the image, producing an OCR text list in which each entry corresponds to a complete readable sentence or phrase rather than fragmented word pieces.

Conditioned on both the image and its OCR text list, the model generates a structured text editing task from one of four predefined categories: \emph{text addition}, \emph{text replacement}, \emph{text removal}, and \emph{hybrid editing}. For each task, it outputs a natural language editing instruction together with the corresponding edited text targets, which are later used as references for normalized edit-distance computation in text accuracy evaluation. We construct 200 samples for each editing type, resulting in a balanced benchmark over the four task categories. To better reflect realistic editing needs and increase task difficulty, a single task may involve multiple edit targets, potentially spanning multiple OCR-recognized sentences rather than a single text instance.

This design ensures that each benchmark sample is both contextually grounded and quantitatively measurable, while establishing a consistent interface between task construction, model generation, and automatic evaluation.

\subsection{Evaluation Metrics}
\label{sec:metric}

We evaluate model performance from three complementary perspectives: \textit{Text Accuracy}, \textit{Visual Quality}, and \textit{Background Preservation}.

\paragraph{Text Accuracy (TA)}
Text editing quality should reflect both successful modification of the target text and faithful preservation of non-target text. Prior evaluation protocols have notable limitations. Retrieval-based methods \cite{du2025textcrafter} typically compare recognized words against an unordered set of reference words, which ignores word order, repetition, and spurious insertions. In contrast, coarse VLM-based scoring \cite{gui2025texteditbench, zhang2026weedit} provides only subjective ratings and lacks sensitivity to fine-grained textual errors.

To address these issues, we measure text accuracy based on the \emph{full-image edit distance}. Given the source image, edited image, and editing instruction, we use GPT-5.2 as the multimodal evaluator to infer the complete text that should appear after editing, including both the edited target text and the text that should remain unchanged. The inferred full text is then compared with the full text actually observed in the edited image. The text accuracy score is defined as
\[
\mathrm{TextAcc} = 1 - \min\left(\frac{S + I + D}{N_{\text{edit}}}, 1\right),
\]
where $S$, $I$, and $D$ denote the numbers of substitutions, insertions, and deletions between the expected and observed full-image text sequences, and $N_{\text{edit}}$ denotes the number of words in the target edit span. We normalize the full-image edit distance by the number of target edited words rather than the total number of words in the image, because the latter is often dominated by preserved background text. Using the full-image text length as the denominator would dilute editing errors and make the metric less sensitive to failures on the actual edit targets. At the same time, because the edit distance is still computed over the entire image text, the metric penalizes both incorrect target edits and unintended corruption of preserved background text.

\paragraph{Visual Quality. (VQ)}
In addition to textual correctness, high-quality text editing should produce edits that are correctly localized, stylistically coherent, and physically plausible. We therefore evaluate visual quality with GPT-5.2 as the multimodal judge along three binary criteria: \textit{location correctness}, which examines whether the edit is applied to the intended region while preserving the position and layout of non-target text; \textit{style consistency}, which evaluates whether the edited text remains compatible with the original visual style and whether preserved text exhibits noticeable style drift; and \textit{physical plausibility}, which assesses readability and visual realism, including consistency in lighting, perspective, boundaries, and local texture. Each criterion is assessed using a yes-or-no judgment rather than a free-form numerical score. This VQA-style formulation reduces the subjectivity of direct scalar rating, mitigates scale ambiguity across samples, and encourages the judge to make explicit decisions on well-defined visual attributes.

\paragraph{Background Preservation (BP)}
Background preservation measures how well non-text visual content is retained after editing. Because textual correctness is already accounted for in \textit{Text Accuracy}, we mask out all detected text regions when evaluating background consistency, preventing reasonable text modifications from being mistakenly penalized as background changes. Specifically, we use PaddleOCR~\cite{cui2025paddleocr30technicalreport} to detect all text bounding boxes in both the source image and the edited image, take their union as an exclusion mask, and further dilate the mask to reduce the influence of residual text boundaries. We then compute SSIM on the remaining non-text pixels, which define the background region. We adopt SSIM as the primary background preservation metric because it better captures perceptual and structural similarity than pixel-wise measures such as MSE or PSNR, making it more suitable for evaluating texture consistency and structural preservation in edited backgrounds.

%% file: sec/5_evaluation.tex
\section{Experiments}
\label{sec:experiments}

This section evaluates TextSculptor against state-of-the-art scene text editing models under the evaluation protocol defined in Sec.~\ref{sec:metric}. We report results from three complementary aspects: \emph{Text Accuracy}, measured by full-image text alignment with edit-distance-based scoring; \emph{Visual Quality}, assessed by GPT-5.2 using binary VQA-style judgments; and \emph{Background Preservation}, computed by SSIM on OCR-masked non-text regions.

\subsection{Evaluation Setups}
Leveraging TextSculpt-Data, we train \textbf{TextSculptor}, a text editing model built upon Qwen-Image-Edit-2511~\cite{wu2025qwen}. We fine-tune the model with LoRA, setting the LoRA rank to 64. Training is conducted for one epoch on 32 GPUs with a learning rate of $1\times10^{-4}$. We use a per-GPU batch size of 4 and set the gradient accumulation steps to 4, resulting in an effective global batch size of 512.

We compare TextSculptor with a diverse set of baselines from three categories. The proprietary models include Gemini-2.5-Flash-Image, Seedream4.0, and Seedream4.5. The open-source general generative models include OmniGen2 \cite{wu2025omnigen2} and Bagel \cite{deng2025emerging}. The open-source editing-oriented models include Step1X-Edit\cite{liu2025step1x}, LongCat-Image-Edit \cite{LongCat-Image}, Qwen-Image-Edit-2511 \cite{wu2025qwen}, and FireRed-Image-Edit-1.0 \cite{team2026firered}. For fair comparison, all baseline methods are evaluated using their default settings. Quantitative results across different text editing types are reported in Table~\ref{tab:main_results_by_type}.

\subsection{Main Results}

\input{tab/main_results}

Table~\ref{tab:main_results_by_type} reports the performance of representative models on TextSculpt-Bench across four text editing types: addition, removal, replacement, and hybrid editing. Overall, existing models exhibit clear trade-offs among Text Accuracy (TA), Visual Quality (VQ), and Background Preservation (BP). Some strong proprietary models, such as Seedream4.5 and Gemini-2.5-Flash-Image, achieve competitive TA and VQ, indicating their ability to generate visually plausible edited images and follow text-related instructions to some extent. However, their BP scores remain relatively limited, suggesting that they may introduce unnecessary changes to non-edited regions during text manipulation. In contrast, several open-source editing models obtain relatively high BP in certain settings but struggle with TA and VQ, showing that preserving the background alone does not guarantee successful text editing. This phenomenon is especially evident in addition and hybrid editing, where models need to simultaneously insert or modify text while maintaining layout consistency and visual realism. Compared with these baselines, TextSculptor achieves a more balanced performance across all three metrics, obtaining the best overall BP score of 0.78 and the best overall average score of 0.69. Notably, TextSculptor performs strongly on removal and replacement tasks, reaching 0.70/0.82/0.77 and 0.74/0.75/0.77 in terms of TA/VQ/BP, respectively. These results demonstrate that our model can perform localized text manipulation while better preserving surrounding image content, which we attribute to the Programmatic Text Editing Data Pipeline that provides explicit supervision for edited and non-edited regions.

\input{tex/vis}

These findings highlight the diagnostic value of TextSculpt-Bench. Unlike general image editing benchmarks that mainly emphasize semantic consistency or overall visual quality, TextSculpt-Bench jointly evaluates three tightly coupled requirements for text editing: text correctness, visual plausibility, and background preservation. The results reveal that text editing cannot be reduced to generic image editing, as it requires models to precisely localize editable text regions, modify textual content according to instructions, and avoid disturbing irrelevant background areas. The clear performance gaps across different editing types and metrics show that TextSculpt-Bench can expose the limitations of existing models and provide a fine-grained evaluation protocol for text editing.


Figure~\ref{fig:vis_metirc} visualizes a evaluation trace for a hybrid text editing case. Given the editing instruction and generated image, our protocol compares the observed text, visual judgments, and background consistency of each edited result. This trace shows how TA captures full-image textual mismatches, VQ decomposes visual quality into location, style, and physical plausibility, and BP measures non-text background preservation. The example also illustrates that TextSculptor better follows the instruction while maintaining comparable background consistency.

\subsection{Ablation Studies}
\label{sec:ablation}

We conduct ablation studies to validate the effectiveness of the proposed data construction strategy. All variants are evaluated on TextSculpt-Bench using the metrics defined in Sec.~\ref{sec:metric}. The results are summarized in Table~\ref{tab:ablation_streams}. We study how different components and design choices in TextSculpt-Data affect the final text editing performance. Besides the baseline model, we compare TextSculptor (Full) with two ablated variants: (1) \textbf{w/o Distraction}, where each Part 2 editing sample contains only one rendered text element pasted onto the background; (2) \textbf{w/o T2I Data}, where the Part 1 text-to-image rendering data is removed from training.

As shown in Table~\ref{tab:ablation_streams}, the full TextSculptor training recipe consistently improves over the baseline, increasing the average score from 0.65 to 0.69. Removing distraction texts leads to a clear drop in Visual Quality, from 0.68 to 0.60, while Background Preservation remains comparable. This suggests that composing multiple text instances onto natural backgrounds provides useful supervision for handling cluttered text layouts and visually complex editing scenarios. Without such distraction patterns, the model is less robust to realistic scenes that contain multiple text regions.

Removing the Part 1 T2I data also degrades performance, especially in Text Accuracy and Visual Quality. The \textbf{w/o T2I Data} variant obtains lower TA and VQ than the full model, indicating that the text-to-image rendering data helps the model learn general text appearance, layout, and integration with visual context.

\begin{table}[h]
\centering
\vspace{-6pt}
\caption{
Ablation on the contribution of T2I rendering data and distraction text composition.
}
\label{tab:ablation_streams}
\renewcommand{\arraystretch}{1.1}
\begin{tabular}{l | ccc | c}
\toprule
\textbf{Data Configuration} & \textbf{TA} $\uparrow$ & \textbf{VQ} $\uparrow$ & \textbf{BP} $\uparrow$ & \textbf{Avg.} $\uparrow$ \\
\midrule
Baseline & 0.55 & 0.63 & 0.76 & 0.65 \\
TextSculptor (Full) & 0.60 & 0.68 & 0.78 & 0.69 \\
w/o Distraction & 0.57 & 0.60 & 0.79 & 0.65 \\
w/o T2I Data & 0.56 & 0.61 & 0.80 & 0.66 \\
\bottomrule
\end{tabular}
\vspace{-6pt}
\end{table}

%% file: tab/main_results.tex
\begin{table*}[t]
\centering
\small
\setlength{\tabcolsep}{3.5pt}
\resizebox{\textwidth}{!}{%
\begin{tabular}{lccc|ccc|ccc|ccc|cccc}
\toprule
\multirow{2}{*}{Model} 
& \multicolumn{3}{c}{Addition} 
& \multicolumn{3}{c}{Removal} 
& \multicolumn{3}{c}{Replacement} 
& \multicolumn{3}{c}{Hybrid} 
& \multicolumn{4}{c}{Overall} \\
\cmidrule(lr){2-4} \cmidrule(lr){5-7} \cmidrule(lr){8-10} \cmidrule(lr){11-13} \cmidrule(lr){14-17}
& TA & VQ & BP 
& TA & VQ & BP 
& TA & VQ & BP 
& TA & VQ & BP 
& TA & VQ & BP & Avg. \\
\midrule
Gemini-2.5-Flash-Image     & 0.52 & 0.58 & 0.72 & 0.61 & 0.72 & 0.71 & 0.73 & 0.73 & 0.71 & 0.68 & 0.73 & 0.72 & 0.63 & 0.69 & 0.72 & 0.68 \\
Seedream4.0            & 0.35 & 0.38 & 0.62 & 0.60 & 0.63 & 0.64 & 0.75 & 0.74 & 0.62 & 0.56 & 0.55 & 0.64 & 0.56 & 0.57 & 0.63 & 0.59 \\
Seedream4.5            & 0.59 & 0.65 & 0.63 & 0.84 & 0.80 & 0.61 & 0.86 & 0.82 & 0.64 & 0.76 & 0.77 & 0.64 & 0.76 & 0.76 & 0.63 & 0.72 \\
\midrule
OmniGen2               & 0.03 & 0.10 & 0.65 & 0.37 & 0.42 & 0.67 & 0.27 & 0.35 & 0.66 & 0.18 & 0.24 & 0.70 & 0.21 & 0.28 & 0.67 & 0.39 \\

Bagel                  & 0.07 & 0.11 & 0.80 & 0.50 & 0.44 & 0.74 & 0.28 & 0.24 & 0.77 & 0.20 & 0.20 & 0.79 & 0.26 & 0.25 & 0.77 & 0.43 \\
Step1X-Edit             & 0.05 & 0.08 & 0.71 & 0.76 & 0.74 & 0.75 & 0.60 & 0.58 & 0.71 & 0.46 & 0.42 & 0.75 & 0.47 & 0.45 & 0.73 & 0.55 \\

LongCat-Image-Edit     & 0.21 & 0.24 & 0.65 & 0.76 & 0.78 & 0.67 & 0.59 & 0.55 & 0.66 & 0.58 & 0.59 & 0.68 & 0.54 & 0.54 & 0.67 & 0.58 \\
FireRed-Image-Edit-1.0 & 0.28 & 0.43 & 0.71 & 0.78 & 0.88 & 0.73 & 0.79 & 0.79 & 0.73 & 0.64 & 0.72 & 0.74 & 0.62 & 0.70 & 0.73 & 0.68 \\
Qwen-Image-Edit-2511   & 0.30 & 0.41 & 0.78 & 0.64 & 0.77 & 0.75 & 0.69 & 0.71 & 0.74 & 0.59 & 0.64 & 0.78 & 0.55 & 0.63 & 0.76 & 0.65 \\

TextSculptor (Ours)    & 0.33 & 0.44 & 0.79 & 0.70 & 0.82 & 0.77 & 0.74 & 0.75 & 0.77 & 0.62 & 0.72 & 0.79 & 0.60 & 0.68 & 0.78 & 0.69 \\
\bottomrule
\end{tabular}
}
\caption{Performance comparison across different text editing types. TA, VQ, and BP denote Text Accuracy, Visual Quality, and Background Preservation, respectively.}
\label{tab:main_results_by_type}
\end{table*}

%% file: tex/vis.tex
\begin{figure*}[t]
\vspace{-6pt}
\begin{center}
   \includegraphics[width=1.0\linewidth]{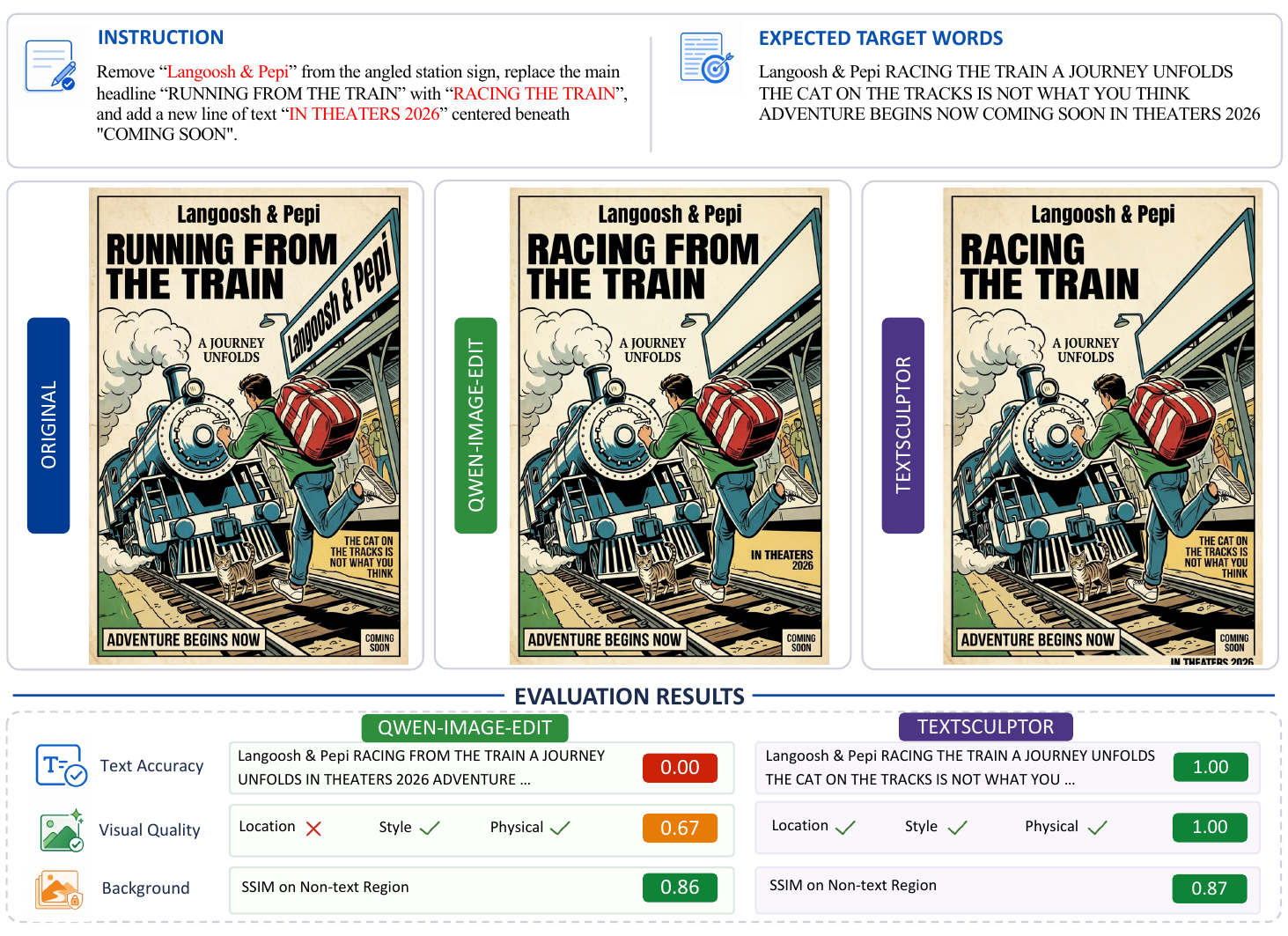}
\end{center}
   \caption{
   Qualitative evaluation example on TextSculpt-Bench.
   }
\label{fig:vis_metirc}
\vspace{-6pt}
\end{figure*}

%% file: sec/6_conclusion.tex
\section{Conclusion}
\label{sec:conclusion}

In this work, we present \textbf{TextSculptor}, a framework that addresses the critical bottlenecks of data scarcity and evaluation in scene text editing. By integrating VLM-based semantic rewriting with programmatic rendering, we constructed \textbf{TextSculpt-Data}, a high-fidelity dataset of 3.2M samples that ensures both textual accuracy and background consistency. We further introduced \textbf{TextSculpt-Bench}, a multi-dimensional benchmark with a tailored protocol for assessing text accuracy, visual quality, and background preservation. Our results demonstrate that TextSculptor narrows the gap to proprietary systems while setting a strong open-source baseline for scene text editing.